\documentclass[letterpaper, 10 pt, conference]{ieeeconf}  

\IEEEoverridecommandlockouts                              

\makeatletter
\let\NAT@parse\undefined
\makeatother
\usepackage[colorlinks,linkcolor=blue,anchorcolor=blue,urlcolor=magenta,citecolor=awesome]{hyperref}

\usepackage{cite}
\usepackage{subcaption}
\usepackage{makecell}
\usepackage{amsmath,amssymb,amsfonts}
\usepackage{algorithmic}
\usepackage{graphicx}
\usepackage{textcomp}
\usepackage{booktabs}
\usepackage{multirow}
\usepackage{xspace}
\usepackage[x11names,table]{xcolor}

\definecolor{awesome}{rgb}{1.0, 0.13, 0.32}

\definecolor{gray}{rgb}{0.3,0.3,0.3}
\definecolor{blue}{rgb}{0,0.5,1}
\definecolor{mask_red}{rgb}{1,0,0.8}
\definecolor{green}{rgb}{0.2,1,0.2}
\definecolor{rblue}{rgb}{0,0,1}
\definecolor{lightblue}{HTML}{6495ed}
\definecolor{lightred}{HTML}{F19C99}

\newcommand{\fn}[1]{\footnotesize{#1}}
\newcommand{\gbf}[1]{\textcolor{orange}{\bf{\fn{#1}}}}

\makeatletter
\let\NAT@parse\undefined
\makeatother

\makeatletter
\DeclareRobustCommand\onedot{\futurelet\@let@token\@onedot}
\def\@onedot{\ifx\@let@token.\else.\null\fi\xspace}
\def\eg{\emph{e.g}\onedot} 
\def\ie{\emph{i.e}\onedot}

\makeatother

\title{\LARGE \bf
\textbf{\textsc{Mate}Robot}: Material Recognition in Wearable Robotics \\ for People with Visual Impairments}

\author{Junwei Zheng$^{1,*}$, Jiaming Zhang$^{1,2,*}$, Kailun Yang$^{3,\dag}$, Kunyu Peng$^{1}$, and Rainer Stiefelhagen$^{1}$
\thanks{This work was supported in part by the Ministry of Science, Research and the Arts of Baden-Württemberg (MWK) through the Cooperative Graduate School Accessibility through AI-based Assistive Technology (KATE) under Grant BW6-03, in part by BMBF through a fellowship within the IFI programme of DAAD, in part by the Helmholtz Association Initiative and Networking Fund on the HAICORE@KIT and HOREKA@KIT partition, and in part by Hangzhou SurImage Technology Company Ltd.
}
\thanks{$^{1}$The authors are with the Institute for Robotics and Anthropomatics, Karlsruhe Institute of Technology, Karlsruhe 76131, Germany.}
\thanks{$^{2}$The author is also with the Department of Engineering Science, University of Oxford, Oxford OX1 3PJ, UK.}
\thanks{$^{3}$The author is with the School of Robotics and the National Engineering Research Center of Robot Visual Perception and Control Technology, Hunan University, Changsha 410082, China.}
\thanks{$^{*}$Equal contribution.}
\thanks{$^{\dag}$Corresponding author: Kailun Yang. (E-mail: kailun.yang@hnu.edu.cn.)}
}

\begin{document}

\maketitle
\thispagestyle{empty}
\pagestyle{empty}

\begin{abstract}
People with Visual Impairments (PVI) typically recognize objects through haptic perception. \emph{Knowing objects and materials before touching} is desired by the target users but under-explored in the field of human-centered robotics. To fill this gap, in this work, a wearable vision-based robotic system, \textsc{Mate}Robot, is established for PVI to recognize materials and object categories beforehand. To address the computational constraints of mobile platforms, we propose a lightweight yet accurate model {\textsc{Mate}ViT} to perform pixel-wise semantic segmentation, simultaneously recognizing both objects and materials. Our methods achieve respective $40.2\%$ and $51.1\%$ of mIoU on COCOStuff-10K and DMS datasets, surpassing the previous method with ${+}5.7\%$ and ${+}7.0\%$ gains. Moreover, on the field test with participants, our wearable system reaches a score of $28$ in the NASA-Task Load Index, indicating low cognitive demands and ease of use. Our \textsc{Mate}Robot demonstrates the feasibility of recognizing material property through visual cues and offers a promising step towards improving the functionality of wearable robots for PVI. The source code has been made publicly available at \href{https://junweizheng93.github.io/publications/MATERobot/MATERobot.html}{MATERobot}.
\end{abstract}

\section{Introduction}

\begin{figure}[t]
    \centering
    \includegraphics[width=0.99\linewidth]{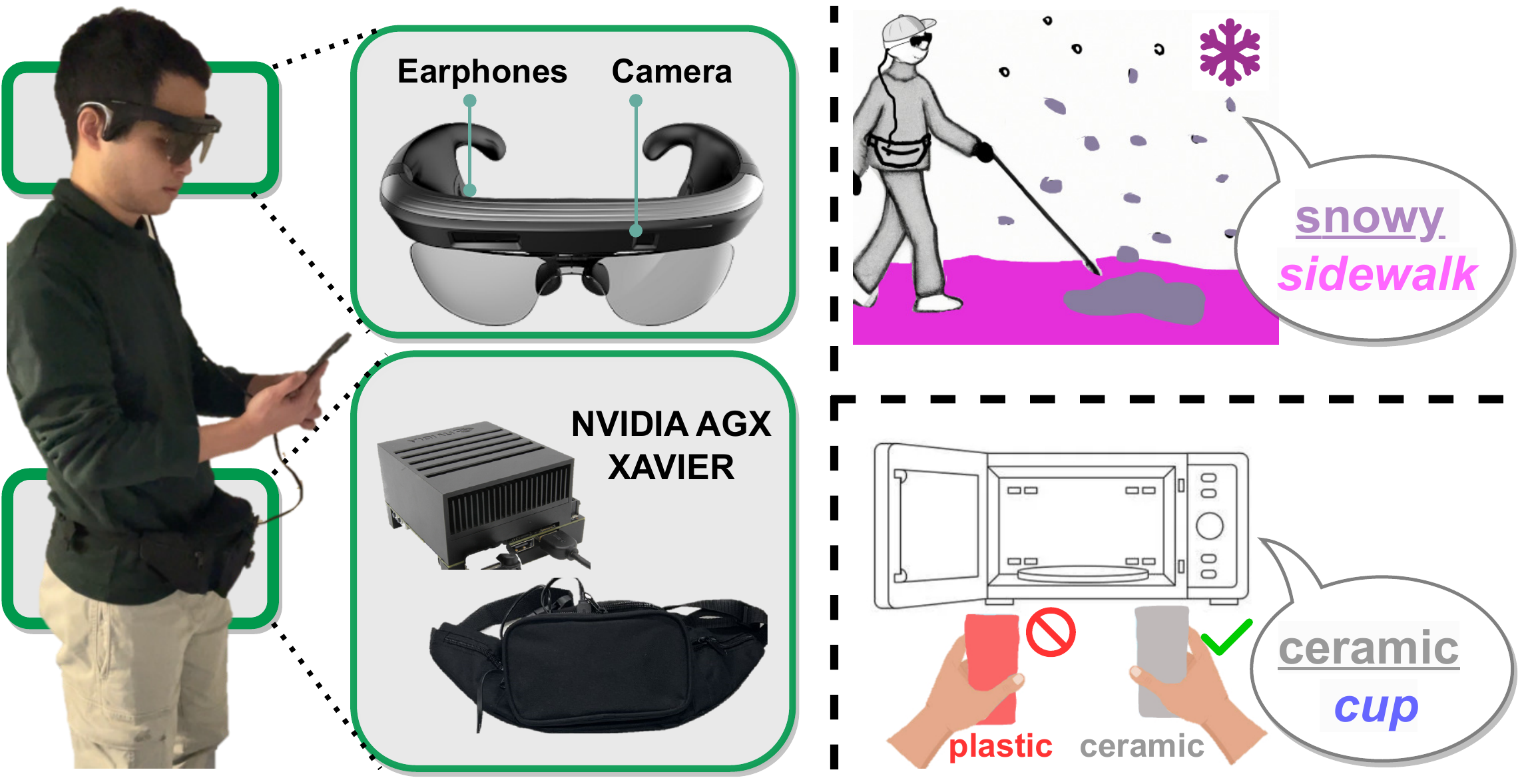}
    \begin{minipage}[t]{.55\columnwidth}
        \vskip-3ex
        \subcaption{Wearable robotic system}\label{fig1_a}
    \end{minipage}%
    \begin{minipage}[t]{.44\columnwidth}
        \vskip-3ex
        \subcaption{Predict \underline{material} \& \textit{object}}\label{fig1_b}
    \end{minipage}%
    \caption{\textsc{Mate}Robot, (a) wearable robotics, can assist (b) \underline{material} semantic segmentation (\eg, \textcolor[HTML]{BD33A4}{\underline{snow}}, \textcolor{gray}{\underline{ceramic}}) and general \textit{object} semantic segmentation (\eg, \textit{\textcolor[HTML]{f013ee}{sidewalk}, \textcolor{blue}{cup}}).}
    \label{fig:head}
\end{figure}

In 2020, it was estimated that approximately 43 million individuals were living with blindness. By 2050, this number is expected to rise significantly to 61 million~\cite{burton2021lancet}. Therefore, it is imperative to facilitate the development of assistive systems for helping PVI. In recent years, considerable progress has been witnessed in the field of human-centered assistive technology, such as systems for navigation~\cite{duh2020v}, object localization~\cite{agrawal2022novel}, indoor understanding~\cite{liu2021hida}, and path orientation~\cite{zhang2022trans4trans}. 

\textit{Material recognition} is often a challenging task for PVI, who typically recognize objects through touch~\cite{paterson2006seeing,klatzky2007object}. However, material recognition is under-explored in the domain of assistive technology.
Therefore, it is essential to develop a wearable system that can assist PVI in recognizing the materials of objects without touching them, \ie, a contact-free material recognition system. 

In this work, 
we present the wearable \textbf{\textsc{Mate}Robot} system to cover both object and material recognition. As shown in Fig.~\ref{fig1_a}, the system consists of a pair of smart glasses with an RGB-D camera, a pair of bone-conducting earphones, and a portable image processor inside a small bag.
In Fig.~\ref{fig1_b}, the {\textsc{Mate}Robot} can recognize \underline{materials} (indicated by underline, \eg, \underline{\textcolor[HTML]{BD33A4}{snow}}) and general \textit{objects} (indicated in italics, \eg, \textit{\textcolor[HTML]{f013ee}{sidewalk}}). These two predictions can form feedback of object categories with material properties (\eg, ``\underline{\textcolor[HTML]{BD33A4}{snowy}} \textit{\textcolor[HTML]{f013ee}{sidewalk}''}). This information is conveyed to the blind user in the form of speech through bone-conduction headphones. %

Due to computationally complex self-attention operations in vision transformers~\cite{dosovitskiy2020ViT}, it is, however, hard to deploy a resource-intensive ViT-based model on portable platforms. To address this, we propose an efficient \underline{mate}rial segmentation \underline{ViT}-based model, namely \textbf{\textsc{Mate}ViT}, which includes a \textit{Learnable Importance Sampling (LIS)} strategy to maintain only the informative tokens for the material segmentation, so as to reduce the computational cost. Thanks to the LIS strategy, a resource-friendly \textsc{Mate}ViT model is obtained, enabling the deployment of vision transformer on 
wearable robotic devices that have limited computational resources.

Apart from the model efficiency, to enlarge the model capacity, we introduce a \textit{Multi-gate Mixture-of-Experts (MMoE)} method to combine the aforementioned general image semantic segmentation and material semantic segmentation on a single model. Compared to the previous method~\cite{zhang2022trans4trans} using a straightforward dual-head structure, our MMoE is used to construct a high-performance and efficient multi-task learning architecture. The feature tokens from input images are forwarded to respective gates and experts to extract informative features for task-relevant decoder heads that generate final semantic segmentation masks.     

In order to endow \textsc{Mate}Robot with robust perception capability, including general object and material semantic segmentation, we perform model training on COCOStuff-10K~\cite{caesar2018coco} and DMS~\cite{upchurch2022dms} datasets, both contain more than $10K$ training samples. Through extensive experiments, our small model obtains $40.2\%$ and $51.1\%$ of mIoU scores, surpassing the previous multi-task learning baseline~\cite{zhang2022trans4trans} by absolute ${+}5.7\%$ and ${+}7.0\%$ on COCOStuff-10K and DMS, respectively. For single-task learning on materials segmentation, \ie, on DMS, our model reaches state-of-the-art performance, having ${+}8.1\%$ mIoU gains compared to previous CNN counterpart~\cite{upchurch2022dms}. To verify the practicability of our \textsc{Mate}Robot for recognizing material categories in real-world scenarios, we conduct a task-oriented user study with six blindfolded participants. On the post-study questionnaires, our system obtains respective $28$ and $77$ scores regarding the evaluation criteria of NASA-Task Load Index (NASA-TLX) and System Usability Scale (SUS), which indicates the ease of use and the usability of our \textsc{Mate}Robot in practical scenarios. 
In summary, our main contributions are:
\begin{itemize}
    \item For the first time, we integrate material recognition into assistive technology and built a wearable robot system, \ie, \textit{\textsc{Mate}Robot}. This system empowers PVI to achieve contactless long-distance perception similar to that of sighted people.
    \item We propose an efficient \textit{\textsc{Mate}ViT} model to enable the deployment of resource-intensive ViT-based counterparts on resource-constrained mobile platforms by using a LIS strategy.
    \item A MMoE method is designed to simultaneously perform object and material semantic segmentation in one unified model. 
    \item Through a task-oriented user study and qualitative analyses, we gain valuable insights into designing wearable material recognition systems for PVI. 
\end{itemize}

\section{Related Work}
\subsection{Wearable Assistive System}
With the tremendous capability revealed by computer vision algorithms, vision-based wearable assistance systems~\cite{yang2018predicting_polarization,wang2017enabling,aladren2014navigation,ou2022indoor,liu2023open} are becoming increasingly applicable. A vision-based navigation system~\cite{duh2020v}, calculating precise positions and orientations, is proposed to help PVI stay on track while walking, and it can recognize unexpected dynamic obstacles. Due to the COVID-19 pandemic, an object-finding algorithm is introduced in~\cite{agrawal2022novel} to build a robotic cane system. 
In~\cite{liu2021hida}, a lightweight system with a solid-state LiDAR sensor is proposed for indoor detection and avoidance. In a previous work~\cite{zhang2022trans4trans,zhang2021trans4trans}, to cover the segmentation of transparent objects, a dual-head model is deployed on a wearable device. 
However, only limited recognizable materials are delivered in previous wearable assistance systems. In order to help blind users obtain a more comprehensive and humanized experience on material recognition, we design a wearable robotic system in this work, for the first time, which can not only recognize conventional object categories, such as \textit{cups}, but also further recognize the material of the object, such as \textit{plastic cups}, delivering contactless object recognition.

\subsection{Material Semantic Segmentation}
Recently, Vision Transformer~\cite{dosovitskiy2020ViT,xie2021segformer} is proposed to utilize the self-attention operation in transformer layers to extract non-local features from a sequence of image patches, yielding an alternative backbone solution compared to convolutional counterparts~\cite{chen2018deeplabv3plus, zhao2017PSPNet, he2016resnet}. In DMS~\cite{upchurch2022dms}, a model based on ResNet~\cite{he2016resnet} is used to address dense material segmentation. In contrast to DMS~\cite{upchurch2022dms}, we adopt Vision Transformer as the backbone, \eg, ViT~\cite{dosovitskiy2020ViT}, to perform material semantic segmentation, with the aim of extracting long-distance dependencies between image patches, since the long-range contextual information is crucial for robust representation of diverse materials~\cite{zhang2022trans4trans}. However, due to the high computational demand of self-attention operation, there is still a bottleneck when deploying a plain vision transformer on resource-constrained mobile platforms, \eg mobile robots and wearable devices. In this work, we propose a novel importance sampling method to reduce the number of tokens and maintain only the informative ones for material segmentation to enable the deployment of plain vision transformers on wearable devices.

\section{\textsc{Mate}Robot: A Wearable Robotic System}

\begin{figure}[t]
    \centering
    \includegraphics[width=0.99\linewidth]{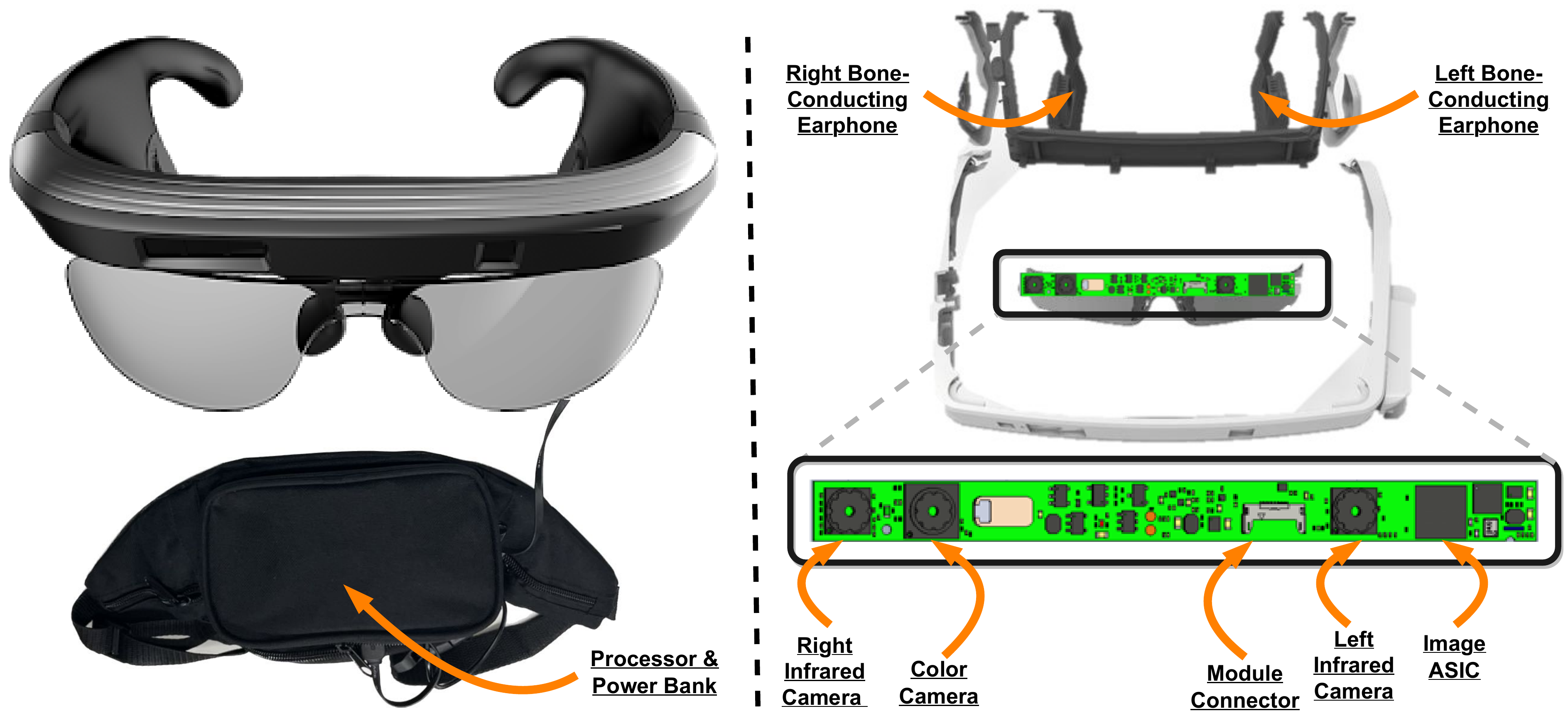}
    \begin{minipage}[t]{.44\columnwidth}
        \vskip-3ex
        \subcaption{Glasses and processor}\label{fig2_a}
    \end{minipage}%
    \begin{minipage}[t]{.48\columnwidth}
        \vskip-3ex
        \subcaption{Components in glasses}\label{fig2_b}
    \end{minipage}%
    \caption{Hardware components in the \textsc{Mate}Robot system.}
    \label{fig2:hardware}
    \vskip-3ex
\end{figure}

\begin{figure*}[t]
    \centering
    \includegraphics[width=1\linewidth
    ]{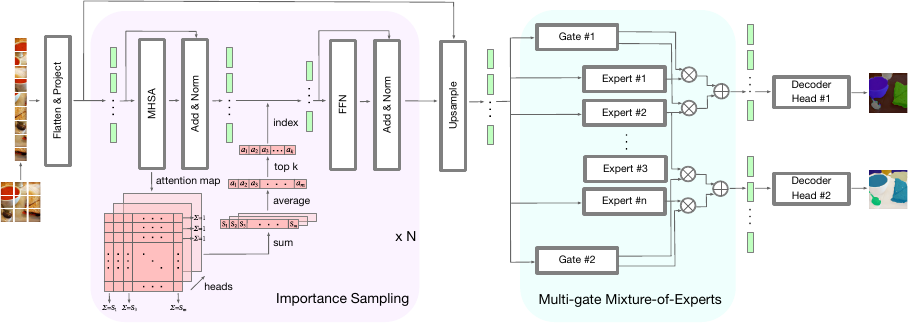}
    \caption{\small \textbf{Architecture of \textsc{Mate}ViT} in a Learnable Importance Sampling (LIS) strategy to reduce the computational complexity, and with a Multi-gate Mixture-of-Experts (MMoE) layer to perform dual-task segmentation (\ie, \#1-Object and \#2-Material segmentation). 
    }
    \label{fig:architecture}
\end{figure*}

\subsection{Hardware Component}\label{method:robots}
As shown in Fig.~\ref{fig2:hardware}, there are three main hardware components in our \textsc{Mate}Robot, including a pair of KRVision smart vision glasses, 
a portable processor, and a power bank inside a waist bag. Inside the smart glasses, there is an RGB-Depth camera RealSense R200 %
and a pair of bone-conduction headphones. For the concept of human-friendly design, there are three advantages of using bone-conduction earphones, which are comfortable to wear, clean and hygienic, and keep in touch with the outside world. Maintaining awareness of ambient sounds is especially important for PVI. To ensure higher portability of the system, we tried different processors and chose the smaller NVIDIA AGX Xavier due to its applicable energy efficiency and inference capabilities. 
Furthermore, a power bank with high energy capacity is selected to provide the system with up to $6$ hours of battery life, which greatly reduces the battery life anxiety of traditional wearable devices~\cite{liu2021hida}. Through the above hardware components, a more portable \textsc{Mate}Robot and better user experience can be delivered to PVI when performing contactless material recognition in real-world scenarios.

\subsection{User Interaction}\label{method:hci}
Between the system and the user, we design an easy-to-use interface for PVI. First of all, in order to give users timely information feedback, we adopt an adjustable feedback frequency. For example, if users set a larger interval in a more familiar environment, \eg, home, they will get concise information. If they explore an unknown space, setting a high frequency can obtain more information. Besides, between the pixel-wise image segmentation results to the auditory output to users, only the detected object and material located in the middle of the input image will be selected. Their pre-defined texts are used to generate speeches jointly in the form of ``\underline{material} \textit{objects}'', such as ``\underline{ceramic} \textit{cups}'' or ``\underline{metal} \textit{forks}''.

\subsection{\textsc{Mate}ViT Model}\label{method:model}
Apart from the hardware components, we further detail the efficient model designed for \textsc{Mate}Robot. 
To equip mobile platforms with plain vision transformers, we propose \textit{\textsc{Mate}ViT}, which has ViT~\cite{dosovitskiy2020ViT} with LIS as the backbone, followed by an upsampling layer, a MMoE layer and two decoder heads. The architecture is shown in Fig.~\ref{fig:architecture}. Note that our method is flexible to include more tasks by adding decoder heads. Mixed by two datasets, each input sample is first encoded by the efficient ViT backbone and then fed into the upsampling layer. One gate in the MMoE layer corresponds to one task, receiving the data sample that is only relevant to the task. Depending on the output of the gate, different selected experts are activated to learn meaningful latent representations, which are finally decoded by the task-relevant decoder head. Different from the training process, one data sample is fed into all gates synchronously during inference, and latent representations from selected experts are then decoded by corresponding decoder heads, producing predictions for all tasks.

\subsection{Learnable Importance Sampling}\label{method:LIS}
To make plain ViT \cite{dosovitskiy2020ViT} more lightweight and feasible in real-world applications, we propose a \textit{Learnable Importance Sampling} strategy for material semantic segmentation, yielding an efficient \textsc{Mate}ViT. Our approach does not require an additional class token in forward pass compared to EViT \cite{liang2022evit}; therefore, it further reduces the model complexity with high-resolution inputs. As illustrated in Fig.~\ref{fig:architecture}, all image patches are flattened and projected into tokens, forming Queries ($\textbf{Q}$), Keys ($\textbf{K}$), and Values ($\textbf{V}$). Multi-Head Self-Attention (MHSA) is then calculated as: 
\begin{equation}
    MHSA(\textbf{Q}, \textbf{K}, \textbf{V})=Softmax(\frac{\textbf{QK}^T}{\sqrt{C}})\textbf{V}
\end{equation}
where $\textbf{Q} {\in}\mathbb{R}^{N \times C}$, $\textbf{K} {\in}\mathbb{R}^{N \times C}$ and $\textbf{V} {\in}\mathbb{R}^{N \times C}$ are query, key and value matrices; $N$, $C$ are token number and dimension. Since softmax is introduced in attention map calculation, the summation of each value in rows is equal to $1$. However, the result does not always equal to $1$ when summing up all values in columns, indicating the importance of each token from an image. Based on this observation, we first calculate the summation in columns and then average the importance vectors among all heads. A fixed number of top $k$ values are selected and tokens are downsampled according to the indices of these $k$ values after \textit{Add \& Norm}, which stand for a residual link \cite{he2016resnet} and layer normalization \cite{ba2016layernorm}. Fig. \ref{fig:architecture} illustrates the whole LIS process in detail. Following \cite{vaswani2017transformer}, the downsampled tokens are then sent to a Feed-Forward Network (FFN) followed by another \textit{Add \& Norm}. Since the token number is reduced after the importance sampling, an upsampling layer is then applied, which is a standard transformer decoder block~\cite{vaswani2017transformer}. Thanks to LIS, a plain ViT-based model is expedited and better qualified for real-world mobile applications.

\subsection{Multi-gate Mixture-of-Experts}\label{method:MMoE}
Since the performance of a wearable robotic system is relevant to its model capacity, Mixture-of-Experts~\cite{shazeer2017moe} is proposed for enlarging model capacity while maintaining invariant model complexity. However, the usage of MoE is less discussed on wearable systems~\cite{zhang2022trans4trans}. For the first time, we adopt the MoE method to perform complementary training of both general object and material segmentation, \ie, the former prevents the latter from overfitting and vice versa. More specifically, we adopt a MMoE layer in our model for high efficiency and performance. Fig.~\ref{fig:architecture} shows the detail of the MMoE layer for multi-task learning. Specifically, one gate is responsible for one task to select the experts. Similar to \cite{shazeer2017moe}, the resulting selection vector $G(\textbf{x})$ can be described as:
\begin{align}
    G(\textbf{x}) &= Softmax(TopM(H(\textbf{x}), m)) \\
    H(\textbf{x}) &= \textbf{x} \cdot \textbf{W}_g + N(\textbf{x})\\
    N(\textbf{x}) &= Normal() \cdot Softplus(\textbf{x} \cdot \textbf{W}_{noise}) \\
    TopM(\textbf{v}, m)_i &= \begin{cases}
                                v_i \hspace{10mm} \text{if $v_i$ ranks top $m$} \\
                                -\infty \hspace{6.5mm} \text{otherwise}
                          \end{cases}
\end{align}
where $\textbf{x}$, $\textbf{W}_g$, and $\textbf{W}_{noise}$ are token, gate matrix, and noise matrix, respectively. The noise term $N(\textbf{x})$ is utilized for load balance, which is the multiplication of the standard normal distribution sampling $Normal()$ and the outcome of $Softplus(\cdot)$. During the training, for every token in one image, only one same gate is activated to produce the selection vector. According to the indices of the top $m$ values, $m$ experts are selected and the token is only fed into the $m$ experts. The output of the MMoE layer is a weighted sum of the top $m$ values in the selection vector and their corresponding outcomes from the $m$ experts. A task-relevant decoder head~\cite{strudel2021segmenter} is then applied to transform all output tokens from MMoE into a prediction mask. Note that we also employ the load and importance balancing loss following~\cite{shazeer2017moe} besides the task loss. During the inference, every token from one image is fed into all gates synchronously. The resulting weighted sums from selected experts are then decoded by the corresponding decoder heads shown in Fig.~\ref{fig:architecture}. As the model is trained jointly on two tasks, the knowledge absorbed from object segmentation plays an important role in the high performance of material recognition, which provides PVI with accurate material information in their daily life.

\section{Experiments}
\subsection{Settings and Datasets}
\noindent\textbf{Settings.} We implement the model with PyTorch $1.12.1$ and CUDA $11.6$. The learning rate is initialized as $0.01$ and it is scheduled by a cosine annealing strategy~\cite{loshchilov2016cosannealing}. SGD with momentum $0.9$ is adopted as the optimizer. We initialize our efficient ViT backbone with a pre-trained plain ViT \cite{dosovitskiy2020ViT} and keep other layers of the model randomly initialized. Data augmentations like random horizontal flipping, random resize with a ratio $0.5\text{-}2$, and random cropping to $512{\times}512$ are used during training. Note that we \textbf{do not} use other tricks such as OHEM, auxiliary loss, and class-weighted loss for a fair comparison to other methods. We train our model with a batch size of $4$ for $200$ epochs on four 1080Ti GPUs.

\noindent\textbf{Datasets.} We adopt COCOStuff-10K \cite{caesar2018coco} and DMS \cite{upchurch2022dms} for general object and material segmentation, respectively. The COCOStuff-10K dataset \cite{caesar2018coco} has $9000/1000$ images for training/testing. We conduct experiments following the implementation of mmsegmentation~\cite{mmseg2020} with $171$ categories. The DMS dataset~\cite{upchurch2022dms} has respective $21857/9057/9152$ images for training/validation/testing with $46$ categories.

\subsection{Results on Material Segmentation}
\begin{figure*}[tbh]
    \centering
    \includegraphics[width=1\linewidth,trim=2 2 2 2,scale=1]{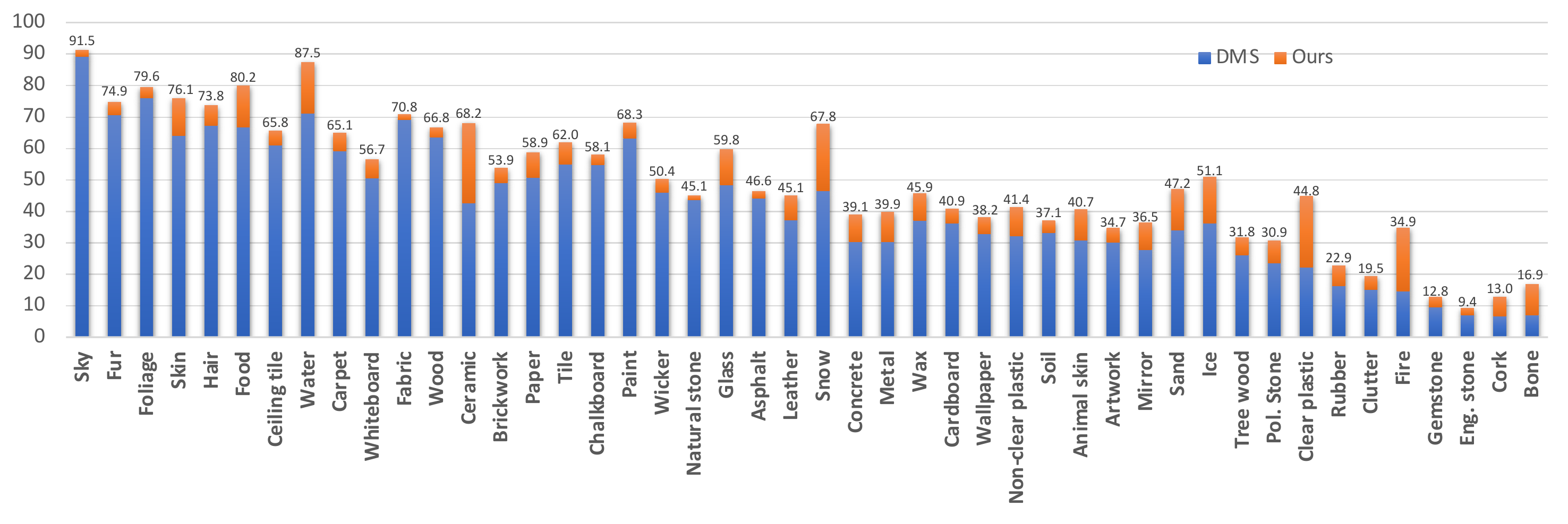}
    \vskip -2ex
    \caption{\textbf{Per-class IoU (\%) of all material categories}. The blue bar shows the category IoU (\%) of the baseline DMS~\cite{upchurch2022dms}, while the \textcolor{orange}{orange} part shows the \textbf{performance gains} (\%) of our proposed method.}
    \label{fig:dms_category_iou}
    \vskip -2ex
\end{figure*}

To verify the proposed method for material segmentation, we conduct experiments of material segmentation on the DMS dataset~\cite{upchurch2022dms}.
Results are reported in Table \ref{tab:sota_dms_camera_ready}. 
It can be observed that our model using the ViT-Tiny backbone still has the lowest computation expense ($4.30$ GFLOPs) with high performance ($44.1\%$ in mIoU), and the ViT-Small variant outperforms other methods in both pixel accuracy and mIoU. Furthermore, Fig.~\ref{fig:dms_category_iou} presents the per-class IoU of all material categories. It is worth noting that our ViT-Small variant achieves performance gains in all $46$ categories, especially those are relevant for assisting PVI, \eg, \textit{fire} (gain ${+}20.2\%$), \textit{snow} (gain ${+}21.3\%$), \textit{plastic} (gain ${+}22.6\%$), and \textit{ceramic} (gain ${+}25.5\%$). 
The impressive pre-study test scores obtained in our evaluations serve as a testament to the effectiveness and reliability of our proposed system in assisting PVI in accurately recognizing and distinguishing materials.
\begin{table}[!t]
\centering
\caption{\textbf{Results of single-task learning on DMS dataset} \texttt{val}/\texttt{test} set. GFLOPs are measured in size of $512 {\times} 512$.}
\label{tab:sota_dms_camera_ready}
\resizebox{\columnwidth}{!}{
\renewcommand{\arraystretch}{1.2}
\setlength{\tabcolsep}{1mm}{
\begin{tabular}{llcc|cc} 
\toprule[1.5pt]
\textbf{Method} & \textbf{Backbone} & \textbf{GFLOPs} & \textbf{MParams} & \textbf{Pixel Acc (\%)} & \textbf{mIoU (\%)} \\ \midrule \midrule
PSPNet \cite{zhao2017PSPNet} & MobileNetV2 & 06.12 & 02.13 & 66.5 / 66.3 & 26.1 / 25.9  \\ 
DeepLabV3 \cite{chen2017deeplabv3} & MobileNetV2 & 07.52 & 02.58 & 68.4 / 68.1 & 29.8 / 29.7  \\ 
DeepLabV3+ \cite{chen2018deeplabv3plus} & MobileNetV2 & 07.85 & 03.11 & 69.3 / 69.2 & 30.1 / 29.9  \\ 
LR-ASPP \cite{Howard2019mobilenetv3} & MobileNetV3-s & 05.41 & \textbf{01.14} & 66.8 / 66.5 & 26.5 / 26.4 \\ 
LR-ASPP \cite{Howard2019mobilenetv3} & MobileNetV3 & 08.78 & 03.29 & 70.7 / 70.2 & 30.3 / 29.9 \\ 
SeaFormer \cite{wan2023seaformer} & SeaFormer-L & 06.50 & 14.00 & 73.1 / 72.8 &  43.4 / 41.8 \\
DMS \cite{upchurch2022dms} & ResNet-50 & - & - & 73.1 / 72.9 &  43.5 / 42.0 \\
\midrule
\rowcolor{gray!20}\textsc{Mate}ViT (ours) & ViT-Tiny & \textbf{04.30} & 07.78 & 76.9 / 76.8 & 45.3 / 44.1 \\ 
\rowcolor{gray!20}\textsc{Mate}ViT (ours) & ViT-Small & 15.54 & {28.79} & \textbf{79.6} / \textbf{79.4} & \textbf{51.0} / \textbf{50.1} \\ 
\rowcolor{gray!10}\textit{w.r.t.} DMS   &   &  &  & \gbf{+6.5} / \gbf{+6.5} &  \gbf{+7.5} / \gbf{+8.1}   \\

\bottomrule[1.5pt]
\end{tabular}
}
}
\end{table}

\subsection{Results on Multi-task Learning}
To deliver more information to PVI and to perform complementary training, we further conduct experiments on multi-task learning based on \textsc{Mate}ViT, covering both object and material segmentation at once. Table \ref{tab:sota_moe} illustrates the quantitative results. Compared to Trans4Trans MiT-B0~\cite{zhang2022trans4trans}, our model with ViT-Tiny requires less computation expense (${-}35.0\%$ GFLOPs), while reaching higher performance on both datasets. Additionally, compared to the Trans4Trans MiT-B2 variant~\cite{zhang2022trans4trans}, it becomes evident that our ViT-Small variant has a higher mIoU on both COCOStuff-10K (gain ${+}5.7\%$) and DMS (gain ${+}7.0\%$).
More important, lower computational complexity (\ie, GFLOPs) is intuitive to reflect the high efficiency of the model running on the mobile platform and, therefore, to improve the user experience of the system. 
Compared to the single-task results in Table~\ref{tab:sota_dms_camera_ready}, multi-task learning brings further improvements. The reason for the gains is two-fold: (1) the MMoE layer enlarges the model capacity; (2) when applying MMoE to both object segmentation and material segmentation, the former prevents the latter from overfitting and vice versa.
\begin{table}[t]
\centering
\caption{\textbf{Results (mIoU) of multi-task learning} on COCOStuff-10k and DMS \texttt{test} sets.
GFLOPs @ $512 {\times} 512$.}
\label{tab:sota_moe}
\setlength{\tabcolsep}{1pt}
\resizebox{\columnwidth}{!}{
\renewcommand{\arraystretch}{1.1}
\setlength{\tabcolsep}{1mm}{
\begin{tabular}{llcc|cc} 
\toprule[1.5pt]
\textbf{Method} & \textbf{Backbone} & \textbf{GFLOPs} & \textbf{MParams} & \textbf{COCOStuff (\%)} & \textbf{DMS (\%)} \\ \midrule \midrule
Trans4Trans~\cite{zhang2022trans4trans} & MiT-B0 & 12.24 & \textbf{04.16} & 27.7 & 37.1 \\
Trans4Trans~\cite{zhang2022trans4trans} & MiT-B1 & 18.98 & 14.26 & 30.6 & 41.3 \\
Trans4Trans~\cite{zhang2022trans4trans} & MiT-B2 & 27.00 & 25.30 & 34.5 & 44.1 \\ \midrule
\rowcolor{gray!20}\textsc{Mate}ViT (ours) & ViT-Tiny & \textbf{07.95} & 11.62 & 32.7 & 45.1 \\
\rowcolor{gray!20}\textsc{Mate}ViT (ours) & ViT-Small & 22.08 & 37.10 & \textbf{40.2} & \textbf{51.1} \\
\rowcolor{gray!10}\textit{w.r.t.} Trans4Trans  & &   &  &  \gbf{+5.7} & \gbf{+7.0}   \\
\bottomrule[1.5pt]
\end{tabular}
}
}
\vskip-3ex
\end{table}

\subsection{Ablation Study}
To fully understand the proposed components, an ablation study is conducted, 
as shown in Table~\ref{tab:ablation_camera_ready}. The baseline model is Segmenter ViT-Tiny~\cite{strudel2021segmenter}. Learned from Table \ref{tab:ablation_camera_ready}, replacing ViT-Tiny with ViT-Small boosts the performance ($43.2\% {\rightarrow}49.2\%$ in mIoU). After applying LIS, the mIoU continuously increases ($49.2\%{\rightarrow}50.1\%$). We then add the MoE layer to our model, leading to an even higher mIoU of $50.7\%$. 
It can be observed that our MMoE model achieves the best performances on all metrics in both object segmentation and material segmentation tasks compared to the baseline model, \ie, $6.5\%$ and $3.7\%$ boosts in pixel accuracy, $8.9\%$ and $7.9\%$ boosts in mIoU. Through the extensive pre-study experiments, the effectiveness of the proposed method can be comprehensively proved. 
\begin{table}[tbh]
\centering
\caption{\textbf{Ablation study} on COCOStuff-10K and DMS. All values are calculated with \texttt{test} sets.}
\label{tab:ablation_camera_ready}
\setlength{\tabcolsep}{4pt}
\renewcommand{\arraystretch}{1.04}
\resizebox{\columnwidth}{!}{
\begin{tabular}{lcccc}
\toprule[1.5pt]
\multirow{2}{*}{\textbf{Method}} & \multicolumn{2}{c}{\textbf{COCOStuff-10K}} & \multicolumn{2}{c}{\textbf{DMS}} \\ \cmidrule(l){2-3} \cmidrule(l){4-5} 
& Pixel Acc (\%) & mIoU (\%) & Pixel Acc (\%) & mIoU (\%)   \\ \midrule \midrule 
Segmenter \cite{strudel2021segmenter} & 65.0 & 31.3 & 76.9 & 43.2 \\ 
\quad + ViT-Small & 69.1 & 38.2 & 79.0 & 49.2 \\
\quad + ViT-Small + LIS & 70.2 & 38.9 & 79.4 & 50.1 \\
\quad + ViT-Small + LIS + MoE & 70.8 & 39.4 & 79.9 & 50.7  \\
\quad + ViT-Small + LIS + MoE + MMoE (\textsc{Mate}ViT) & \textbf{71.5} & \textbf{40.2} & \textbf{80.6} & \textbf{51.1} \\  \bottomrule[1.5pt]
\end{tabular}}
\vskip -2ex
\end{table}

\subsection{Qualitative Analysis}
\begin{figure}[tbh]
    \centering
    \captionsetup[subfigure]{labelformat=empty}
    \includegraphics[width=0.99\linewidth]{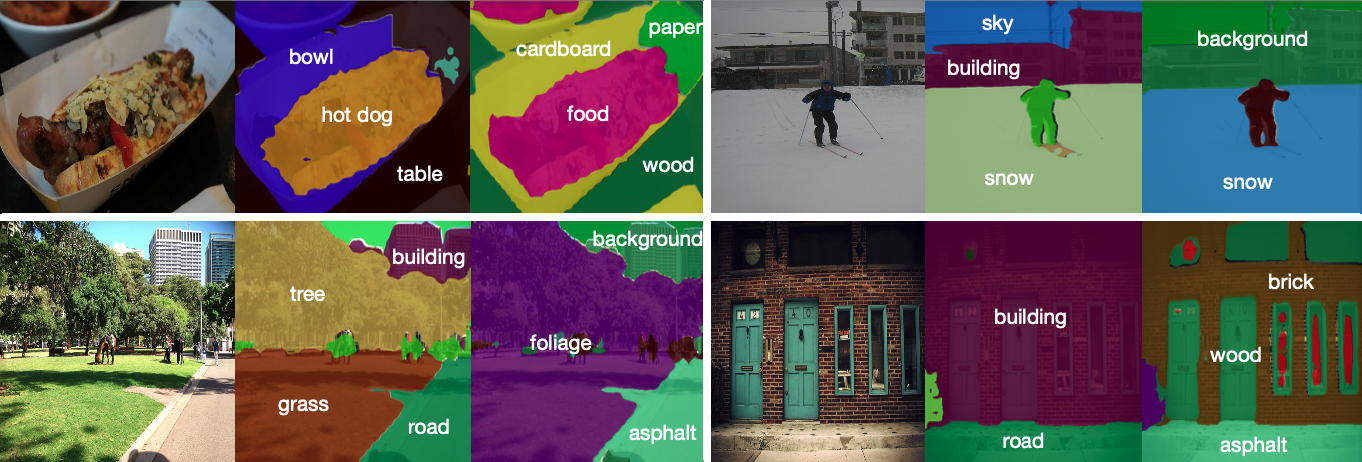}
    \begin{minipage}[t]{.16\linewidth}
        \vskip-3ex
        \subcaption[]{Input}
    \end{minipage}%
    \begin{minipage}[t]{.16\linewidth}
        \vskip-3ex
        \subcaption{Object}
    \end{minipage}%
    \begin{minipage}[t]{.16\linewidth}
        \vskip-3ex
        \subcaption{Material}
    \end{minipage}%
    \begin{minipage}[t]{.16\linewidth}
        \vskip-3ex
        \subcaption{Input}
    \end{minipage}%
    \begin{minipage}[t]{.16\linewidth}
        \vskip-3ex
        \subcaption{Object}
    \end{minipage}%
    \begin{minipage}[t]{.16\linewidth}
        \vskip-3ex
        \subcaption{Material}
    \end{minipage}%
    \caption{\small \textbf{Visualization} of both general object segmentation and material segmentation. From left to right in each group: RGB input, object segmentation, and material segmentation.}
    \label{fig:seg_results}   
    \vskip-2ex
\end{figure}

Four groups of scenarios related to the daily life of PVI are shown in Fig. \ref{fig:seg_results}. 
The upper-left group describes a scenario where PVI are having their meals. The \textit{hot dog} colored orange is perfectly recognized by our system in the second image, and it is tagged with a \underline{food} property from the third image. The group in the lower left corner shows a scenario where blind people walk in a park. According to the predictions in the second and third images, blind people using our system know additionally there is an \underline{asphaltic} \textit{road} ahead. 
In the bottom-right case, the entrance is not recognized if only object recognition is performed, however, the material segmentation result can provide supplementary information to find the doors, which can further improve the mobility accessibility. As a result, our system can help PVI better understand their surroundings, and to support them to make correct interactions with the environment.

\section{User Study}
Following detailed pre-study experiments based on datasets, for wearable robotic systems, a critical factor that needs to be addressed is whether the system can provide a positive user experience in real-world situations. To know about that, we conduct a user study in a structured manner, including task-oriented testing and a questionnaire session.

\subsection{Organization}
To verify the system's usability, we organize a user study with six blindfolded participants in real-world testing scenarios. We select seven categories of materials that are common in daily life, \ie, \textit{fabric}, \textit{foliage}, \textit{glass}, \textit{metal}, \textit{paper}, \textit{plastic}, and \textit{wood}. To conduct a comparison between without and with using our system, there are two rounds of material recognition, \ie, \textbf{contact} and \textbf{contactless} round. The contact round is to recognize by touch, while the contactless round is to recognize by using our system. Note that, to conduct a fair comparison, the order of the objects in two rounds is randomized. 
The participants are required to name the material after touching an object. The reaction time from touching the object until naming the material is recorded by the organizer. 
After two rounds of material recognition, all participants take part in an anonymous questionnaire session. The questionnaires regarding the NASA-Task Load Index (NASA-TLX) and System Usability Scale (SUS) are filled out by all participants. Besides, there is space for participants to write down their open comments.
\begin{figure}[!t]
    \centering
    \includegraphics[width=0.99\columnwidth]{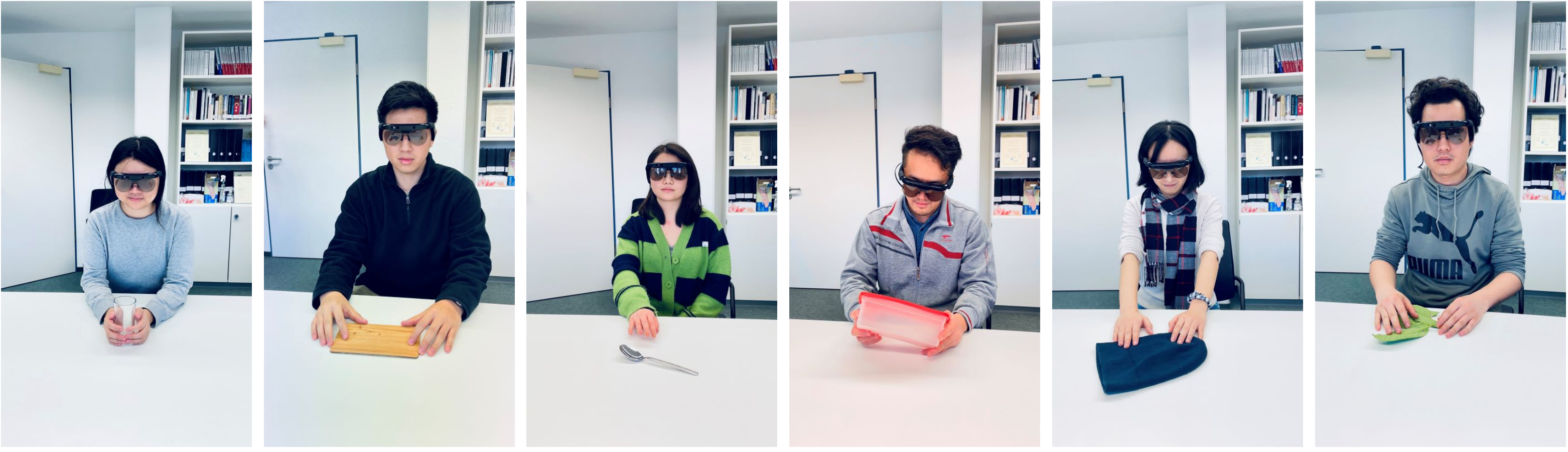}
    \vskip-1ex
    \caption{Incidences of participants on the user study.}
    \label{fig:participants}
    \vskip-4ex
\end{figure}

\subsection{Results and Discussion}

\begin{figure}[tbh]
    \centering
    \includegraphics[width=0.7\linewidth,scale=1]{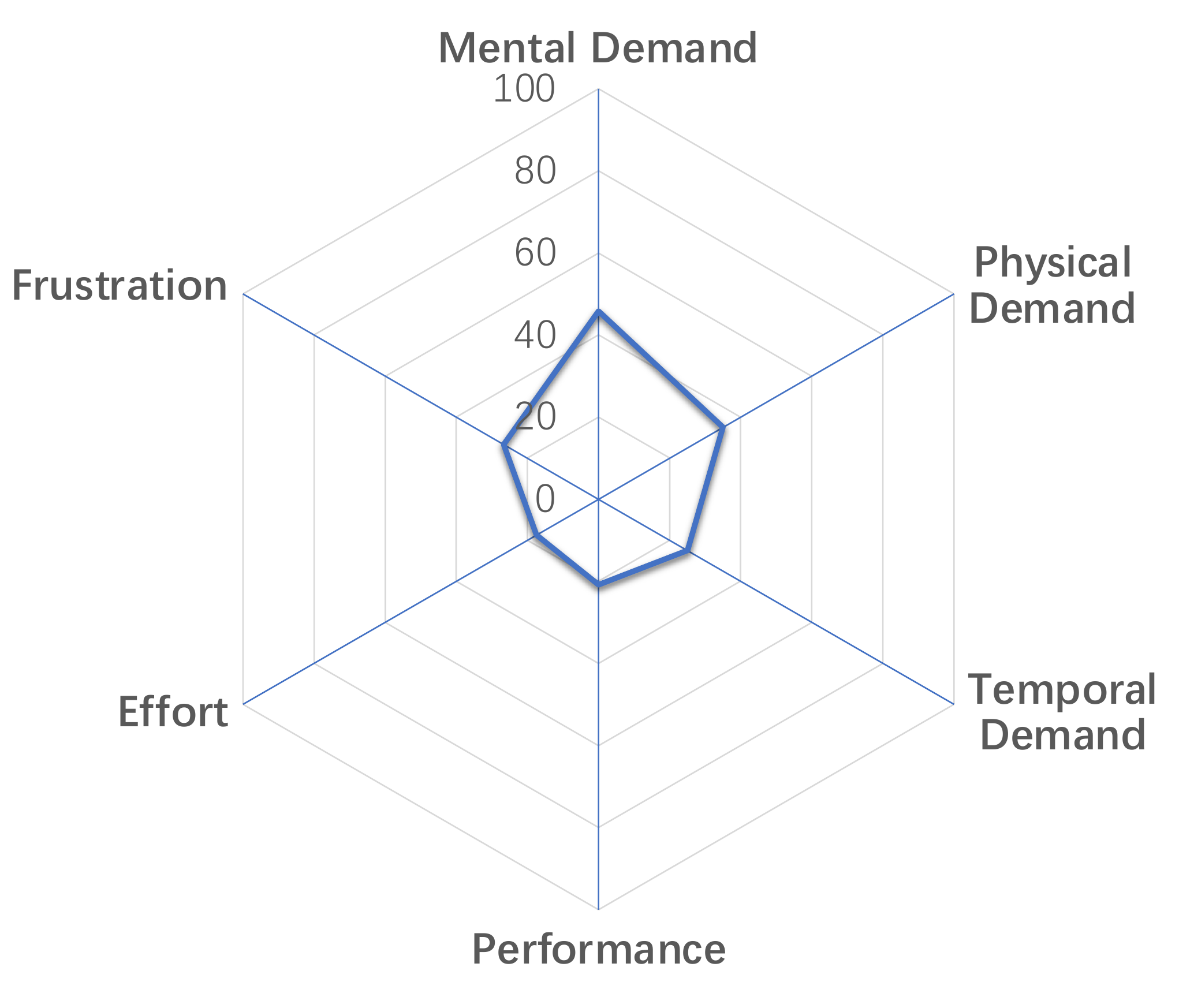}
    \vskip-1ex
    \caption{\textbf{Average NASA-TLX score} of every factor among all participants. Values range from $0$ to $100$, lower is better. Our system requires low cognitive load, with a score of $28$.}
    \label{fig:tlx}
\end{figure}

\noindent\textbf{Cognitive load.}
To learn about the cognitive load of our wearable system, NASA-TLX is a simple and effective method for cognitive load measurement. We first calculate the average score of every factor among all participants, then average the scores of all six factors, resulting in a final NASA-TLX value of $28$. According to \cite{grier2015tlx}, this value illustrates the workload caused by our system is in the 20th percentile of global workload scores from $6.21$ to $88.50$ among $1173$ observations, which can assist users without too much burden. From Fig. \ref{fig:tlx}, we notice that the effort value is relatively smaller than the rest five factors, meaning that our system is user-friendly.

\noindent\textbf{Accuracy.} The correctness of the recognition is defined as the accuracy. In both rounds of material recognition, the accuracy achieves $98\%$, indicating our system is useful and valuable in real-world applications.

\noindent\textbf{Efficiency.}
The time from touching an object to naming the material is defined as the reaction time. We utilize the average reaction time of all materials to evaluate the efficiency. The average reaction time in the contactless round is ${3.11}$s (\textpm $0.21$), while the contact one is $3.97$s (\textpm $0.34$). Without any haptic perception, our wearable system can perform a faster recognition than the contact-based perception, which indicates our \textsc{Mate}Robot is feasible and reliable to provide fast assistance in recognizing material properties.
More importantly, contactless object and material recognition can provide psychological safety for PVI by allowing them to identify objects without touching them.

\noindent\textbf{Usability.}
Apart from NASA-TLX, we verify the usability of our system with SUS. Our system scores $77$ out of $100$, which is a relatively high score. According to Bangor~\textit{et al.}~\cite{bangor2008sus}, who analyzed $2324$ surveys from $206$ studies, ``\textit{the best quarter of studies range from $78.51$ to $93.93$}''. Therefore, we find that our system is useful for recognizing not only general objects but also their material properties.

\noindent\textbf{User comments.}
We analyze the open comments made by users during the testing and from the post-study questionnaire session. The insights are reported below: (1) $66.7\%$ participants were amazed by the fast response of the \textsc{Mate}Robot system, which is one of the reasons why they would like to use the system. (2) The user experience with the system was impressive for all participants. They found our system useful and helpful in the daily lives of visually impaired people. (3) Some participants suggest that the voice feedback of the glasses should constantly inform the user of the detected objects and materials. We plan to improve this in future work. 

\noindent\textbf{Limitations.}
Due to the challenges posed by the pandemic, it was difficult to find a sufficient number of participants to perform a user study on the developed system. The system was tested by six blindfolded participants. While the field test provides insights into the ability of our system, the results cannot be considered representative of the experience of real blind users since PVI could be better at contact perception than sighted people. 
For future work, we plan to conduct studies involving users who are blind or visually impaired. 
%

\section{Conclusion}
In this work, we look into semantic material understanding for helping visually impaired people via a wearable robotic system \textsc{Mate}Robot. We put forward \textsc{Mate}ViT, which unifies general object and material segmentation via an MMoE architecture, whose efficiency is enhanced via LIS to make plain-ViT models suitable for mobile applications. The proposed model is ported to our established assistive \textsc{Mate}Robot system designed for supporting PVI. Extensive experiments on DMS and COCOStuff-10K datasets and a user study demonstrate the effectiveness and usefulness of our recognition system.


\clearpage
\bibliographystyle{IEEEtran}
\bibliography{bib}

\end{document}